\documentclass[letterpaper]{article} 
\usepackage[preprint]{aaai2027}  
\usepackage[hyphens]{url}  
\usepackage{graphicx} 
\usepackage{natbib}  
\usepackage{caption} 
\usepackage{algorithm}
\usepackage{algorithmic}

\usepackage{newfloat}
\usepackage{listings}
\DeclareCaptionStyle{ruled}{labelfont=normalfont,labelsep=colon,strut=off} 
\floatstyle{ruled}
\newfloat{listing}{tb}{lst}{}
\floatname{listing}{Listing}

\usepackage{booktabs}

\usepackage{booktabs}
\usepackage{array} 
\usepackage{multirow}
\usepackage{colortbl}
\usepackage{xcolor}
\usepackage{amsmath} 
\definecolor{promptframe}{RGB}{64,64,64}
\definecolor{promptback}{RGB}{243,243,243}
\lstdefinestyle{promptbox}{
    basicstyle=\footnotesize\ttfamily,
    backgroundcolor=\color{promptback},
    rulecolor=\color{promptframe},
    frame=single,
    frameround=tttt,
    framesep=5pt,
    xleftmargin=4pt,
    xrightmargin=4pt,
    numbers=none,
    breaklines=true,
    breakatwhitespace=false,
    columns=fullflexible,
    keepspaces=true,
    showstringspaces=false,
    aboveskip=8pt,
    belowskip=8pt,
    captionpos=t
}

\title{Agent-R$^2$: Scaling Agents for Realistic Productivity Work via Real-Source\\ Data Synthesis}
\author{
    Teng Pan\textsuperscript{\rm 1,\rm 2},
    Zixuan Wang\textsuperscript{\rm 1,\rm 2},
    Yuchen Yan\textsuperscript{\rm 1},
    Qifei Wu\textsuperscript{\rm 2},
    Yao Chen\textsuperscript{\rm 2},\\
    Yingqi Qu\textsuperscript{\rm 2},
    Ruiqing Zhang\textsuperscript{\rm 2},
    Yongliang Shen\textsuperscript{\rm 1,\ensuremath{\dagger}}
}
\affiliations{
    \textsuperscript{\rm 1}Zhejiang University, \textsuperscript{\rm 2}Baidu Inc.
}

\begin{document}

\maketitle
{\renewcommand{\thefootnote}{}\footnotetext{Code is available at \url{https://github.com/ZJU-REAL/Agent-R2}}}

\begin{abstract}
Large language models have made rapid progress in agentic reasoning and tool use, raising expectations for agents that can take over realistic productivity work from humans. However, current agents still struggle to finish such tasks, where success requires following long-horizon professional workflows, understanding heterogeneous file contexts, and producing usable deliverables under fine-grained requirements. A key bottleneck is the lack of scalable post-training data that reflects real work intent, authentic file environments, and deliverable-level evaluation. To close this gap, we introduce Agent-R$^2$, a $\textbf{r}$eal-source data synthesis framework that scales training data for $\textbf{r}$ealistic productivity work. Instead of relying on manually specified workflows or LLM-generated files, Agent-R$^2$ grounds synthesis throughout the pipeline in real occupational seeds, corpus-derived workflow evidence, and authentic web-sourced files. From 10k web-corpus candidates, Agent-R$^2$ distills 4k task instances that pair real multi-format attachments with fine-grained, deliverable-level rubrics. On this data, we train Agent-R$^2$-4B via supervised fine-tuning and rubric-based reinforcement learning, improving both end-to-end productivity performance and file-grounded reasoning. Trained on only 4k instances, Agent-R$^2$-4B surpasses Qwen3-235B-A22B-Thinking-2507 on both GDPval (51.0$\%$ vs. 48.0$\%$) and OfficeQA (14.8$\%$ vs. 13.8$\%$). Further analysis shows that Agent-R$^2$ mitigates \emph{shortcut behavior}, where agents produce superficially plausible deliverables before sufficiently grounding, constructing, or verifying them, while promoting sustained exploration.
\end{abstract}


\section{Introduction}
\label{sec:introduction}
Large language models have made remarkable progress in agentic reasoning and tool use, enabling agents to plan, interact with environments, and solve increasingly open-ended tasks~\citep{yao2023reacta,schick2023toolformer,zhou2024webarena}. These advances raise the expectation that agents can take over \emph{realistic productivity work} from humans. Such work is \textbf{economically valuable but demanding}: it involves long-horizon professional workflows, where agents must understand professional contexts, gather and integrate evidence from heterogeneous digital artifacts, invoke the necessary tools, and ultimately deliver usable work products such as analytical reports, spreadsheets, presentations, and structured documents.


However, current agents still fall short on such work. First, real professional workflows involve interdependent steps, implicit domain conventions, and evolving intermediate states, so models miss critical operations or fail to adapt based on feedback. Second, real file contexts are heterogeneous and messy: models must integrate evidence across artifacts with different formats, layouts, and levels of completeness, and often misread content or overlook cross-file dependencies. Third, real deliverables must satisfy fine-grained requirements over factual consistency, evidence coverage, structure, formatting, and practical usability; producing an output file is not the same as meeting deliverable standards. Mastering all three demands at once is precisely what current training data fails to teach.

\begin{figure}[t]
    \centering
    \includegraphics[width=\linewidth]{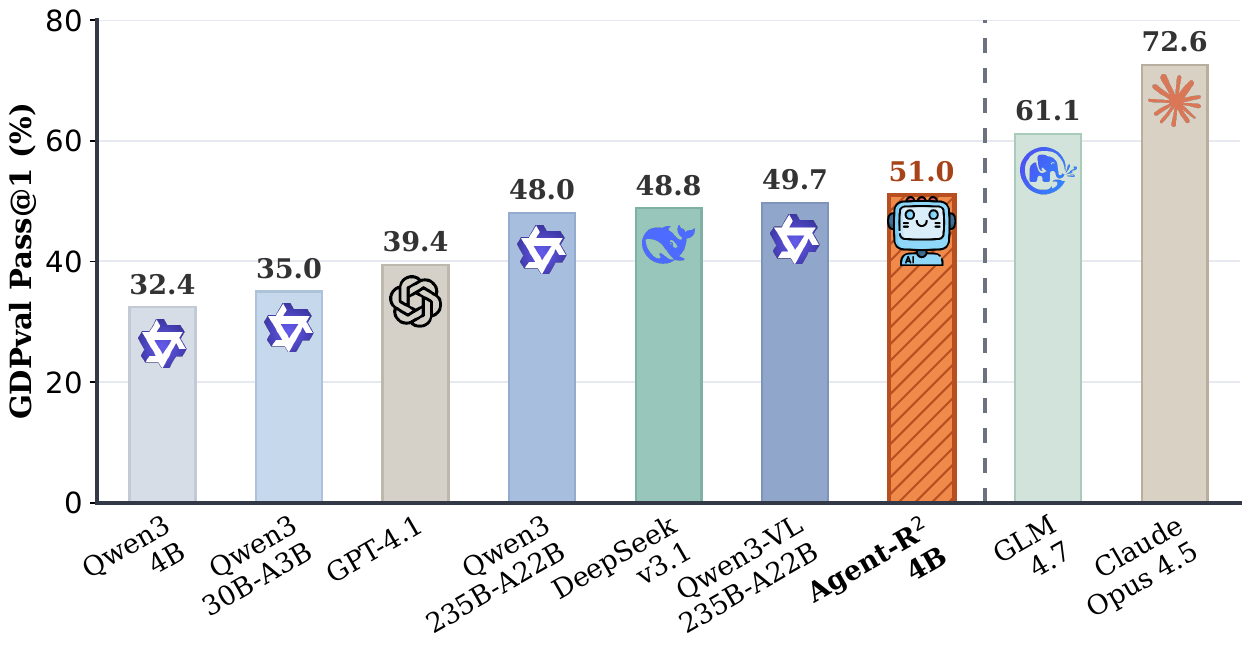}
    \caption{\textbf{GDPval Pass@1 comparison.} Agent-R$^2$-4B improves the Qwen3-4B-Thinking base model by 18.6 points and surpasses several larger open-source baselines.}
    \label{fig:intro_gdpval_pass1}
\end{figure}

Existing training data targets these capabilities only in isolation: instruction following~\citep{xu2025wizardlm,guo2025recast}, multidisciplinary reasoning~\citep{liu2025designer}, document-grounded question answering~\citep{tanaka2023slidevqa,zhang2026docdancera}, or generation of a single deliverable type such as presentations and spreadsheets~\citep{zheng2025pptagent,zheng2026deeppresenter,chi2026spreadsheetrl}. While effective in their respective settings, these data sources do not provide end-to-end supervision that jointly covers real work objectives, cross-file evidence, professional workflow execution, and deliverable constraints. \textbf{What is missing, we argue, is not more data but data that reflects real work: real intents, authentic files, and deliverable-level standards.}

Such data is hard to fabricate: manually authored workflows do not scale, and LLM-generated files lack the layouts, gaps, and cross-file references of real documents. To address this data bottleneck, we introduce Agent-R$^2$, a \emph{real-source} data synthesis framework that grounds every stage of synthesis in real sources through a three-stage pipeline. First, rather than relying on manually authored workflows as in recent benchmark construction~\citep{patwardhan2025gdpvalb,chen2026agentifoneday}, Agent-R$^2$ \textbf{mines file-oriented workflow blueprints} from web corpora using occupational activity seeds, improving both realism and scalability. Second, it \textbf{searches for and verifies public files} that support each workflow, retaining the heterogeneous formats, layouts, incomplete structures, and cross-file dependencies of realistic digital work. Third, it \textbf{synthesizes user instructions and fine-grained rubrics} from the verified workflow-file contexts: the instructions require complex deliverables involving evidence gathering, reasoning, file operations, and artifact construction, while the rubrics specify detailed criteria for evidence grounding, analytical correctness, and deliverable quality. Starting from 10k web-corpus candidates, Agent-R$^2$ produces \textbf{4k high-quality training instances}.

We train Agent-R$^2$-4B on the synthesized data using supervised fine-tuning followed by rubric-based reinforcement learning. The reinforcement learning stage optimizes the model in the persistent sandbox with outcome rewards: each generated deliverable is evaluated against its fine-grained rubric, which encourages the model to satisfy evidence, reasoning, and artifact-quality requirements rather than merely produce an output file. Trained on only 4k synthesized instances, Agent-R$^2$-4B reaches \textbf{51.0\%} Pass@1 on the full GDPval set (Figure~\ref{fig:intro_gdpval_pass1}), improving the base model by \textbf{18.6 points} and surpassing Qwen3-235B-A22B-Thinking-2507 (48.0\%) and Qwen3-VL-235B-A22B-Thinking (49.7\%).


The gains transfer to document-grounded reasoning on OfficeQA: Agent-R$^2$-4B improves the base model from 2.2\% to 14.8\% Pass@1 and again surpasses Qwen3-235B-A22B-Thinking-2507 (13.8\%). Behavioral analysis further shows that training on rubric-scored real-source data mitigates \emph{shortcut behavior}, a general failure mode where agents emit superficially plausible deliverables before sufficiently grounding, constructing, or verifying them, while promoting sustained exploration.

Our contributions are summarized as follows:

\begin{itemize}
  \item We introduce Agent-R$^2$, a data synthesis framework that grounds workflow extraction, file collection, and instruction-rubric synthesis in real sources, scaling training data for realistic productivity work.
  \item We synthesize \textbf{4k} high-quality productivity-work instances with authentic multi-format attachments and fine-grained rubrics, and train Agent-R$^2$-4B with SFT and rubric-based RL.
  \item Agent-R$^2$-4B surpasses Qwen3-235B-A22B-Thinking-2507 on GDPval and OfficeQA; further analysis indicates that the training mitigates \emph{shortcut behavior} while promoting sustained exploration.
\end{itemize}

\section{Problem Formulation}
\label{sec:problem_formulation}

We study realistic productivity tasks, in which an agent must carry out long-horizon professional work in a sandbox environment with general tools, heterogeneous input files, and deliverable-level requirements.

\paragraph{Task.}
A task instance is denoted as
\[
x=(I,\mathcal{F}_{a},\mathcal{F}_{r},\mathcal{R}),
\]
where \(I\) is the user instruction, \(\mathcal{F}_{a}\) is the attachment set available during solving, \(\mathcal{F}_{r}\) is an optional reference set used only for evaluation, and \(\mathcal{R}\) is the task-specific rubric. When no reference files are provided, \(\mathcal{F}_{r}=\emptyset\). The instruction \(I\) specifies the required deliverables, which the agent produces as a set of generated files \(\mathcal{F}_{d}\).

\paragraph{Environment.}
Both solving and evaluation use a persistent sandbox environment
\[
\mathcal{E}=(\mathcal{S},\mathcal{T},\delta,\Omega),
\]
where \(\mathcal{S}\) is the state space, \(\mathcal{T}\) is the tool set, \(\delta\) is the transition function, and \(\Omega\) is the observation function. The tool set is
\[
\mathcal{T}=\{\textit{bash},\textit{web\_search},\textit{web\_fetch},\textit{view\_image}\}.
\]
The \textit{bash} tool executes shell commands and file operations inside the sandbox. The \textit{web\_search} and \textit{web\_fetch} tools are used to locate relevant webpages and retrieve the content of a specified webpage, respectively. The \textit{view\_image} tool provides visual access to images or rendered file pages, and its return depends on whether the solver has multimodal capability. If so, the visual observation is returned directly; otherwise, the tool requires a textual visual query \(q\) and invokes an auxiliary vision policy \(\pi_{v}\) to return the result:
\[
o^{v} =
\left\{
\begin{array}{ll}
m, & \mbox{with multimodal capability},\\
\pi_{v}(m,q), & \mbox{without multimodal capability},
\end{array}
\right.
\]
where \(o^{v}\) denotes the tool's return, \(m\) is an image, and \(q\) is the textual visual query used only when an auxiliary vision policy is needed.

\paragraph{Solving and Evaluation.}
During solving, the sandbox is initialized with the attachment files \(\mathcal{F}_{a}\). The solver's goal is to generate the deliverable files \(\mathcal{F}_{d}\) requested by \(I\). At turn \(t\), the solver receives an observation, produces an action conditioned on the instruction and the interaction history, and the environment transitions accordingly:
\[
o_t=\Omega(s_t), \quad a_t\sim\pi_s(\cdot\mid I, o_{\le t}, a_{<t}), \quad s_{t+1}=\delta(s_t,a_t).
\]
Evaluation uses a separate sandbox initialized with the generated deliverable files \(\mathcal{F}_{d}\) and, when available, the reference files \(\mathcal{F}_{r}\). The evaluator policy \(\pi_e\) interacts with \(\mathcal{E}\) to judge the deliverables under the rubric \(\mathcal{R}\) and outputs the final score.

\section{Data Synthesis}
\label{sec:data_synthesis}

\subsection{Pipeline}
\label{subsec:data_synthesis_pipeline}

\begin{figure*}[t!]
    \centering
    \includegraphics[width=\linewidth]{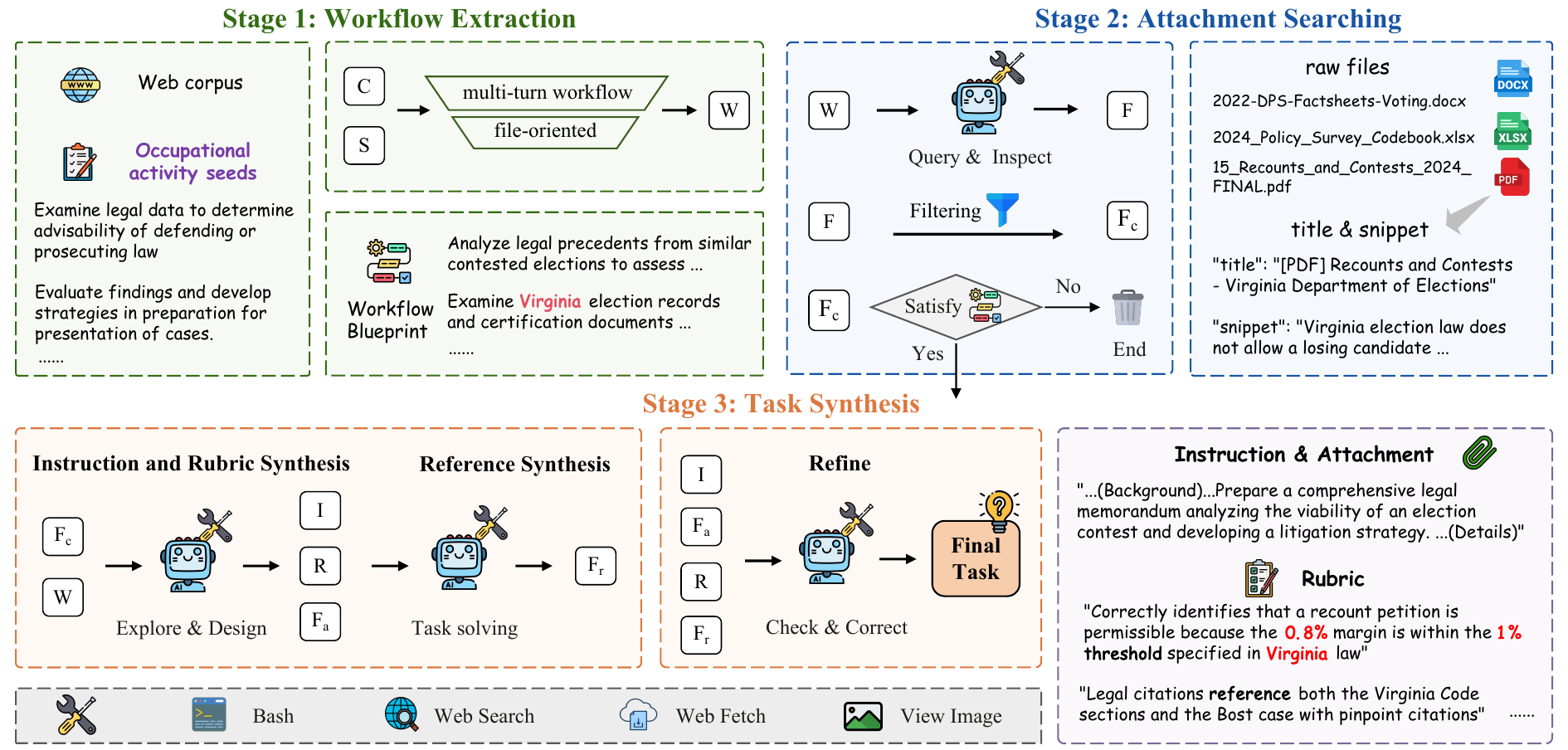}
    \caption{\textbf{Agent-R$^2$ data synthesis pipeline.} The framework extracts file-oriented workflow blueprints from real web-corpus evidence, retrieves and verifies public attachments, and synthesizes instructions, rubrics, and references in sandbox.}
    \label{fig:pipeline}
\end{figure*}

In this section, we detail the data synthesis pipeline of Agent-R$^2$. The overall pipeline is illustrated in Figure~\ref{fig:pipeline}. The prompts used in each stage are provided in the supplementary material.

\paragraph{Workflow Extraction.}
\label{para:workflow_extraction}
Prior work observes that raw web text can encode implicit tool-use workflows~\citep{xu2026unlocking}. Inspired by this observation, we further treat web corpora as a natural source of workflow evidence for realistic productivity work and focus on file-oriented workflows, where completing the work requires locating, reading, editing, or producing file artifacts. By leveraging broad web corpora, Agent-R$^2$ enables scalable workflow construction and avoids the limited scalability of manually curated workflows~\citep{patwardhan2025gdpvalb,chen2026agentifoneday}. We begin with a web corpus \(\mathcal{C}\) and occupational activity seeds \(\mathcal{G}\). These seeds can be drawn from different occupational resources; in this work, we use seeds from O*NET. For the web corpus, we use samples from the DLR-Web dataset~\citep{attention1115dlrweb} used by DESIGNER~\citep{liu2025designer}, which are filtered from FineFineWeb~\citep{map2025finefineweb}. Each segment has corpus-provided discipline metadata, which we map to an occupational category to select the corresponding seeds from \(\mathcal{G}\). For each segment, we perform workflow extraction under two filtering criteria: workflow filtering and file-oriented filtering. A segment that satisfies both criteria is converted into a workflow blueprint, information dependencies, required file types, and expected outputs, where the extraction is performed by Qwen3.6-Plus as the synthesis model \(\pi_{\text{syn}}\):
\[
\mathcal{W}=\{w \mid c\in\operatorname{Filtered}(\mathcal{C}),\;
w\sim\pi_{\text{syn}}(\cdot\mid c,\mathcal{G})\}.
\]

\paragraph{Attachment Searching.}
\label{para:attachment_searching}
Given the workflow blueprints \(\mathcal{W}\), we retrieve public file artifacts and construct candidate file sets.
Unlike prior settings built on large amounts of LLM-generated synthetic files~\citep{ge2026synthetic}, we ground each task in authentic files retrieved from the web, guided by cues that the extracted workflows provide; such files preserve the natural complexity of file structures, layouts, and cross-file evidence.
For each workflow \(w\), we use \(\pi_{\text{syn}}\) to run a multi-turn search-and-verification loop. At each turn, it generates a search query conditioned on the workflow, required file types, previous queries, and prior search results, and executes \(\textit{web\_search}\) to obtain results that include titles and snippets. For each result, it first checks its title and snippet for workflow relevance; promising candidates are then downloaded and inspected with \(\textit{bash}\) and \(\textit{view\_image}\) to check key parts such as cover pages and tables of contents. Files verified as relevant to the workflow are retained in \(\mathcal{F}_c\); to control download latency and construction overhead, we discard files longer than 50 pages or larger than 2 MB. The loop stops when the retained files jointly satisfy the information and operation needs of \(w\), or the maximum number of search turns is reached. Only satisfied pairs are passed to the next stage, while unsatisfied instances are discarded.
\[
\begin{aligned}
\mathcal{F}_c&=\{f\mid \mathrm{rel}(f,w)=1\},\\
\mathcal{P}&=\{(w,\mathcal{F}_c)\mid
w\in\mathcal{W},\; \mathrm{sat}(w,\mathcal{F}_c)=1\}.
\end{aligned}
\]

\begin{figure*}[t!]
    \centering
    \includegraphics[width=\linewidth]{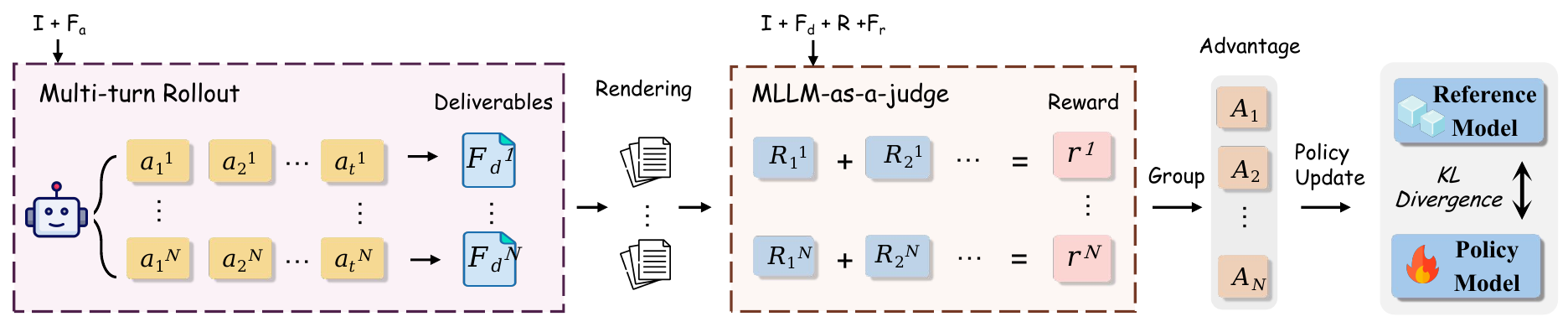}
    \caption{\textbf{Rubric-based RL training.} Agent-R$^2$ samples multi-turn rollouts in sandbox, renders generated deliverables for MLLM-as-a-judge scoring against rubrics and references, and updates the policy with group advantages and KL regularization.}
    \label{fig:train}
\end{figure*}

\paragraph{Task Synthesis.}
\label{para:instruction_synthesis}
We instantiate each retained workflow and candidate-file pair \((w,\mathcal{F}_c)\in\mathcal{P}\) inside the sandbox environment \(\mathcal{E}\). To expose candidate-file content to the model, we render paginated files into page images and parse spreadsheets into sheet-level tables while preserving the original files for inspection. Conditioned on the workflow blueprint and the workflow-relevant candidate files, \(\pi_{\text{syn}}\) performs a multi-turn synthesis process. Across turns, it inspects file contents, searches and fetches web evidence when the workflow requires online information, determines which candidate files should be included as the attachment set \(\mathcal{F}_{a}\subseteq\mathcal{F}_c\), and jointly designs the instruction \(I\) and rubric \(\mathcal{R}\). The instruction specifies the expected deliverable, while the rubric ties success criteria to file-level evidence and constraints.

We then use Claude-Opus-4.5 as the expert policy \(\pi_{\text{exp}}\) to solve the synthesized task in the sandbox. The resulting files are kept as reference files \(\mathcal{F}_{r}\) for evaluation and judging. Along with the deliverable, we also record the expert trajectory \(\tau\), which captures the tool interactions used to inspect files, execute operations, and construct the final output. Let \(u=(I,\mathcal{F}_{a},\mathcal{R})\) denote the candidate solving context:
\[
(\tau,\mathcal{F}_{r})\sim\pi_{\text{exp}}(\cdot\mid u;\mathcal{E}).
\]
We further verify each synthesized candidate by evaluating the generated reference files against the rubric. The verifier identifies rubric items that are not satisfied by \(\mathcal{F}_{r}\). For each unmet item, if the item is unsupported by the attachments, mismatched with the instruction, or otherwise unreasonable, we revise the rubric; if the item is reasonable, we keep the rubric and ask the expert policy to revise the deliverables once in the sandbox. The retained instances form the final task set \(\mathcal{D}=\{(I,\mathcal{F}_{a},\mathcal{F}_{r},\mathcal{R})\}\) and the corresponding verified expert trajectories \(\mathcal{Z}\).

\subsection{Statistics}
\label{subsec:data_synthesis_statistics}
Starting from 10k corpus samples, Agent-R$^2$ yields 4k realistic productivity task instances. Table~\ref{tab:attachment_profile} summarizes the real web-sourced office attachments. The type distribution follows real-source availability, with more PDFs and fewer presentations, and thereby preserves the authenticity of file environments across occupations. The average sizes and page counts further show that these files retain practical content, layout, and format complexity. Figure~\ref{fig:data_statistics} further summarizes task-level characteristics. Each task contains 2.51 attachments on average, requiring agents to inspect and integrate evidence from multiple artifacts (Figure~\ref{fig:data_statistics}(a)). The rubrics are fine-grained and comprehensive, covering both content and formatting requirements, with 15.3 items on average (Figure~\ref{fig:data_statistics}(b)), indicating that tasks require multiple localized reasoning and computation steps as well as both explicit and latent instruction following. Finally, Figure~\ref{fig:data_statistics}(c) reflects expert-trajectory complexity: trajectories contain 18.0 steps and 41.8k tokens on average, indicating that solving the tasks requires long-horizon evidence gathering, file manipulation, and deliverable construction.

\section{Experiment}
\label{sec:experiment}

\subsection{Training Framework}
\label{sec:training_framework}

\begin{table*}[h]
\centering
\begin{tabular*}{\linewidth}{
  @{\extracolsep{\fill}}
  >{\raggedright\arraybackslash}l
  *{6}{c}
  @{}
}
\toprule
\multirow{2}{*}{\textbf{Model}}
& \multicolumn{2}{c}{\textbf{Office}}
& \multicolumn{2}{c}{\textbf{Non-Office}}
& \multicolumn{2}{c}{\textbf{Full}} \\
\cmidrule(lr){2-3} \cmidrule(lr){4-5} \cmidrule(lr){6-7}
& Pass@1 & Pass@3 & Pass@1 & Pass@3 & Pass@1 & Pass@3 \\
\midrule
\rowcolor{gray!10} \multicolumn{7}{l}{\textit{\textbf{Closed-Source Models}}} \\
GPT-4.1~\citep{openai2025gpt41}                      & 38.0 & 51.5 & 44.9 & 55.5 & 39.4 & 52.3 \\
Gemini-2.5-Pro~\citep{comanici2025gemini}               & 57.0 & 59.4 & 61.4 & 63.3 & 57.9 & 60.2 \\
Claude-Opus-4.5~\citep{anthropic2025claudeopus45}              & 73.8 & 77.4 & 68.2 & 72.3 & 72.6 & 76.3 \\
\midrule
\rowcolor{gray!10} \multicolumn{7}{l}{\textit{\textbf{Open-Source Models}}} \\
DeepSeek-v3.1~\citep{deepseek-ai2025deepseekv3}  & 48.3 & 57.5 & 50.9 & 61.7 & 48.8 & 58.4 \\
GLM-4.7~\citep{team2025glm45} & 60.8 & 67.1 & 61.9 & 68.3 & 61.1 & 67.4 \\
Qwen3-30B-A3B-Thinking-2507~\citep{yang2025qwen3a}  & 33.7 & 38.5 & 40.4 & 44.9 & 35.0 & 39.8 \\
Qwen3-235B-A22B-Thinking-2507~\citep{yang2025qwen3a} & 46.9 & 52.9 & 52.0 & 58.6 & 48.0 & 54.1 \\
Qwen3-VL-235B-A22B-Thinking~\citep{bai2025qwen3vl}  & 48.4 & 50.9 & 54.7 & 57.9 & 49.7 & 52.4 \\
\midrule
\rowcolor{gray!10} \multicolumn{7}{l}{\textit{\textbf{Agent-R$^2$ (Ours)}}} \\
Qwen3-4B-Thinking-2507  & 31.1 & 35.7 & 37.8 & 42.3 & 32.4 & 37.0 \\
Agent-R$^2$-4B-SFT   & 45.9 & 51.5 & 42.8 & 47.6 & 45.3 & 50.7 \\
\textbf{Agent-R$^2$-4B}
                             & \textbf{51.5} & \textbf{54.7}
                             & \textbf{48.9} & \textbf{55.1}
                             & \textbf{51.0} & \textbf{54.8} \\
\bottomrule
\end{tabular*}
\caption{\textbf{Main results on GDPval.} Pass@$k$ allows up to $k$ attempts and scores the first attempt that produces a deliverable; results are reported for office, non-office, and full subsets. Agent-R$^2$-4B consistently improves over its base and SFT variants and outperforms several larger open-source baselines on the full set.}
\label{tab:main_gdpval_results}
\end{table*}

Using the synthesized task set \(\mathcal{D}\) and filtered expert trajectories \(\mathcal{Z}\), we train Qwen3-4B-Thinking-2507 in two stages: supervised fine-tuning and rubric-based reinforcement learning~\citep{huang2025reinforcement,gunjal2025rubrics}. SFT teaches the model to follow expert trajectories, inspect files, and perform tool-based operations for deliverable construction. RL further optimizes the policy for rubric-scored outcome quality, encouraging error recovery and stronger satisfaction of task requirements. We split the data into 3k instances for SFT and 1k tasks for RL, with no task overlap.

\paragraph{Rubric-Based Reinforcement Learning.}
Starting from the SFT model, we optimize the policy with GRPO~\citep{shao2024deepseekmatha} in the persistent sandbox environment, as illustrated in Figure~\ref{fig:train}. Since agent-as-a-judge evaluation requires additional interactive tool use and is inefficient during RL training, we use Qwen3.6-Plus as an MLLM-as-a-judge~\citep{chen2024mllmasajudge} for outcome scoring. After task completion, the generated deliverable, relevant file evidence, and reference files are converted into model-consumable representations: .pdf, .pptx, and .docx files are rendered into images, while .xlsx files are represented by recalculated values and formulas in Markdown together with rendered page images, preserving both cell-level content and visual layout. The judge then scores the result against the synthesized rubric and reference files. To avoid inducing tool-specific calling preferences, we use only outcome reward and do not add explicit penalties for intermediate tool-call errors. The reward is computed as
\[
r =
\begin{cases}
0, & \text{if } \hat{\mathcal{F}}_{d}=\emptyset,\\
\frac{\operatorname{Score}(\hat{\mathcal{F}}_{d},\mathcal{F}_{r},\mathcal{R})}
{\text{max score}}, & \text{otherwise},
\end{cases}
\]
where \(\hat{\mathcal{F}}_{d}\), \(\mathcal{F}_{r}\), and \(\mathcal{R}\) denote the generated deliverable files, synthesized reference files, and task-specific rubric used by the judge, respectively.

\subsection{Experiment Setup}
\label{sec:experiment_setup}

\paragraph{Baselines.}
We compare our proposed models against two primary categories of baselines: closed-source models, including GPT-4.1, Gemini-2.5-Pro, and Claude-Opus-4.5, and open-source models, including DeepSeek-v3.1, GLM-4.7, and Qwen3-series 30B and 235B text-only and multimodal variants. For Qwen-series text-only models, \textit{view\_image} calls are handled by the same-series multimodal model with a matching parameter scale. For DeepSeek-v3.1, we use DeepSeek-v4-Flash for \textit{view\_image} calls.

\paragraph{Benchmarks.}
We evaluate model performance on both in-domain and out-of-domain benchmarks. GDPval~\citep{patwardhan2025gdpvalb} serves as the in-domain benchmark, and we further categorize its tasks by the required output deliverable type into \textit{office} and \textit{non-office} tasks. The latter covers outputs beyond office formats, including audio (.wav, .mp3), video (.mp4), and other media or file types, which are outside our primary focus in this work. OfficeQA~\citep{opsahl-ong2026officeqa} serves as the out-of-domain benchmark. We use its official Markdown version, where office source files are provided as plain-text Markdown files, to evaluate whether training improves document perception and reasoning rather than only the surface quality of generated deliverables.

\paragraph{Evaluation Protocol.}
For GDPval, we evaluate deliverables against GDPval rubrics with an agent-as-a-judge
protocol~\citep{openai2025gdpvaldataset}, using Gemini-3.1-Pro as the judge model~\citep{google2026gemini31pro}.
For OfficeQA, we combine rule-based matching with model-based assessment. Both benchmarks set a maximum of 100 steps. We report aggregate benchmark performance to quantify the incremental contributions of the SFT and RL stages.

\subsection{Main Results}
\label{sec:main_result}

\paragraph{Agent-R$^2$ Delivers Strong Performance.}
Table~\ref{tab:main_gdpval_results} shows GDPval results. On the full set, Agent-R$^2$-4B surpasses Qwen3-235B-A22B-Thinking-2507 (51.0\% vs. 48.0\%) despite using a much smaller backbone, and also outperforms DeepSeek-v3.1 and GPT-4.1. The advantage is especially meaningful on office tasks, our primary focus in this work, where Agent-R$^2$-4B comes within 4.7 points of Gemini-2.5-Pro on Pass@3. Both SFT and RL consistently improve the base model on office and non-office tasks. More specifically, SFT brings larger gains on office tasks by imitating expert trajectories, while RL contributes larger gains on the less directly targeted non-office tasks, highlighting its importance for generalization beyond specific formats.

\begin{table}[h]
\centering
\setlength{\tabcolsep}{4pt}
\renewcommand{\arraystretch}{1.08}
\begin{tabular}{lcc}
\toprule
\textbf{Model} & \textbf{Pass@1} & \textbf{Pass@3} \\
\midrule
Qwen3-4B-Thinking-2507 & 2.2 & 4.5 \\
Qwen3-30B-A3B-Thinking-2507 & 2.9 & 5.7 \\
Qwen3-235B-A22B-Thinking-2507 & 13.8 & 25.6 \\
\midrule
Agent-R$^2$-4B-SFT & 11.4 & 21.7 \\
Agent-R$^2$-4B & \textbf{14.8} & \textbf{26.8} \\
\bottomrule
\end{tabular}
\caption{\textbf{OOD results on OfficeQA.} Pass@$k$ measures whether at least one of $k$ attempts yields a correct answer ($k=1,3$). Agent-R$^2$-4B outperforms Qwen3-235B-A22B-Thinking-2507.}
\label{tab:officeqa}
\end{table}

\paragraph{Agent-R$^2$ Generalizes Beyond Productivity Work.}
Table~\ref{tab:officeqa} shows that the improvement transfers to OfficeQA. Both SFT and RL improve over the base model, and the final Agent-R$^2$-4B further surpasses Qwen3-235B-A22B-Thinking-2507 on both Pass@1 and Pass@3. These results suggest that training on synthesized productivity work improves not only deliverable construction, but also the underlying ability to inspect files, integrate evidence, and extract numerical answers across documents.

\subsection{Analysis}
\label{sec:analysis}

We further analyze how SFT and RL reshape the model's behavior, and show that our synthesized data help mitigate shortcut behavior through grounded exploration and deliverable verification.

\begin{figure}[t]
\centering
\includegraphics[width=\linewidth]{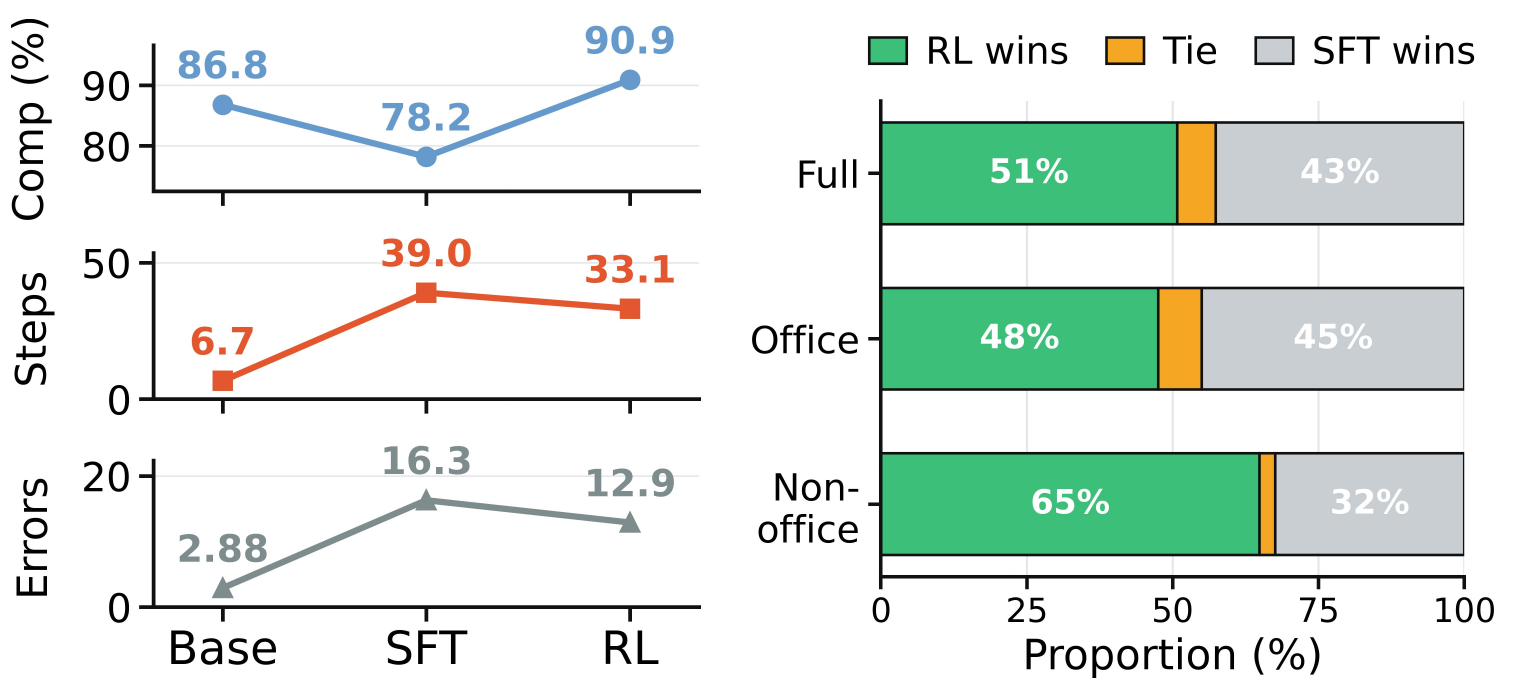}
\caption{\textbf{Model behavior and completion quality.} Left: completion rate, average tool-call steps, and average tool-call errors across the base, SFT, and RL variants. Right: win/tie/loss outcomes of the RL-trained model against the SFT-only model after valid completion on GDPval, where green, orange, and gray segments indicate RL wins, ties, and SFT wins, respectively.}
\label{fig:sft_rl_behavior_tooluse}
\end{figure}

\paragraph{SFT Encourages Sustained Exploration.}
The left panel of Figure~\ref{fig:sft_rl_behavior_tooluse} shows that SFT substantially increases the model's exploratory behavior. The base model makes only 6.7 tool-call steps and 2.88 tool-call errors per sample on average. Such limited interaction suggests that the base model often stops active exploration after encountering early parsing, execution, or file-handling errors, and instead rushes to construct deliverables with insufficient grounding. Agent-R$^2$-4B-SFT increases the average number of tool-call steps to 39.0, with tool-call errors also increasing to 16.3. This shift indicates that SFT teaches the model to inspect files more actively, execute multi-step workflows, and continue attempting corrections after errors. Although limited model capacity and realistic environmental complexity lead to more errors, the longer exploration process ultimately guides the model toward more complete deliverables. Due to limited generalization, not every exploration attempt produces a successful deliverable, as reflected by the completion-rate drop from 86.8\% to 78.2\%, which limits the potential gain. However, this shapes a more proactive behavior pattern and lays the foundation for subsequent RL optimization.


\paragraph{RL Improves Reliable and High-Quality Completion.}
RL preserves the exploratory behavior introduced by SFT and turns it into more reliable task completion. Figure~\ref{fig:sft_rl_behavior_tooluse}(left) shows that Agent-R$^2$-4B raises the completion rate to 90.9\%, above both the base and SFT variants. On the one hand, it averages 33.1 tool-call steps, far above the base model's 6.7, indicating that it does not fall back to sparse inspection and premature output. On the other hand, tool-call errors decrease from 16.3 under SFT to 12.9, suggesting that rubric-based RL helps the model recover from errors, explore more efficiently, and generalize better in complex environments. Furthermore, the completion quality improves. Figure~\ref{fig:sft_rl_behavior_tooluse}(right) compares Agent-R$^2$-4B with the SFT-only variant on GDPval tasks that both models complete successfully. The comparison shows that RL wins more often than SFT on both office and non-office tasks, indicating that rubric-based RL improves not only completion rate but also the quality of deliverables.



\begin{figure}[t]
\centering
\includegraphics[width=\linewidth]{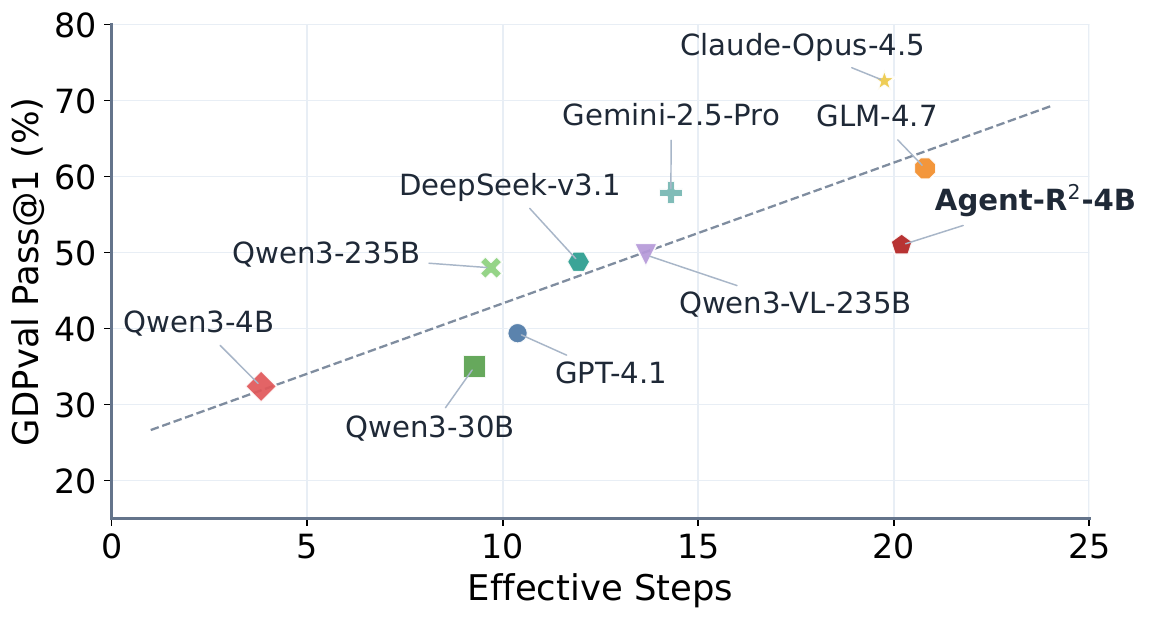}
\caption{\textbf{Effective Steps vs. GDPval Performance.} Effective steps are computed as average tool-call
steps minus tool-call errors. Models with more effective steps generally achieve higher GDPval Pass@1.}
\label{fig:effective_steps_pass1}
\end{figure}

\begin{figure*}[t!]
    \centering
    \includegraphics[width=\linewidth]{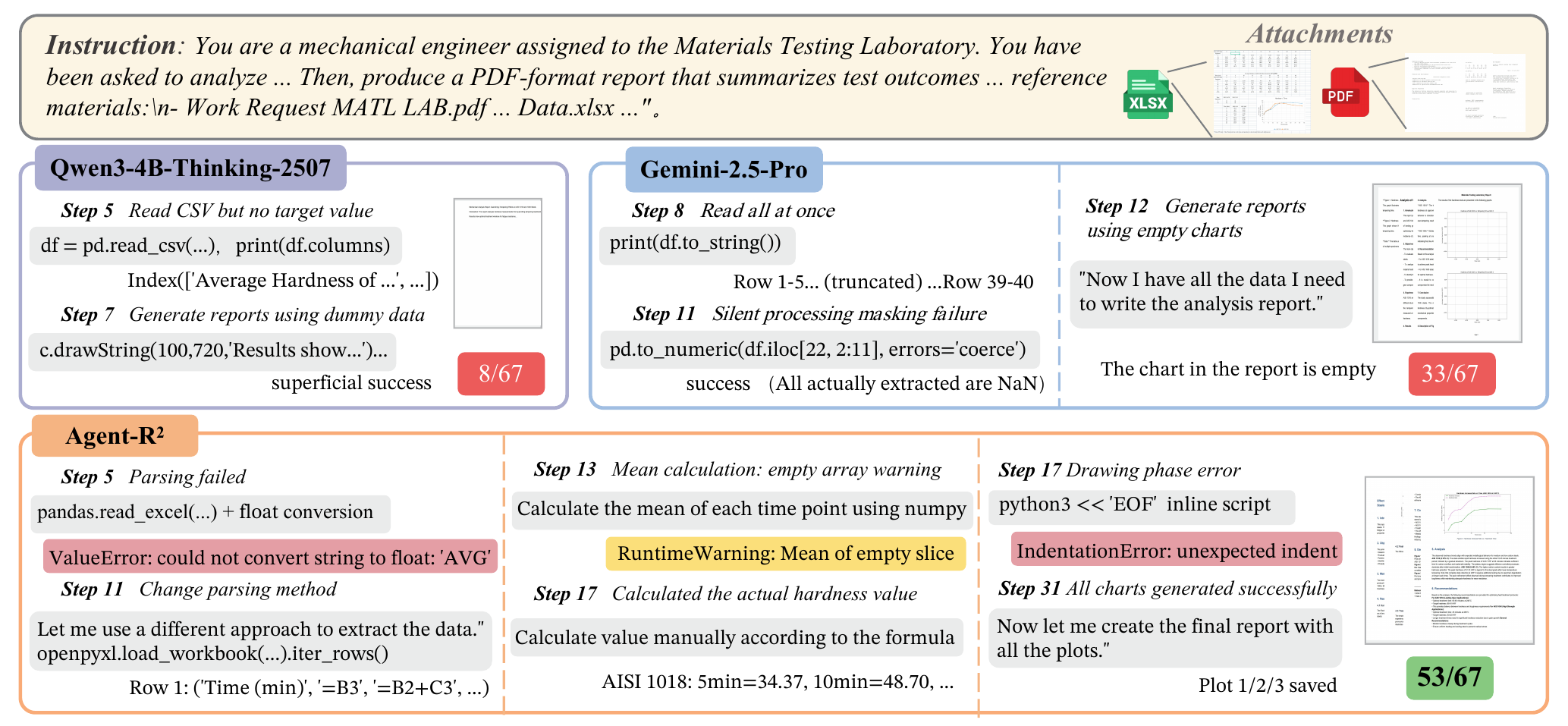}
    \caption{\textbf{Case study of shortcut behavior in materials-test report generation.} The base model and Gemini-2.5-Pro produce flawed reports from incomplete or invalid spreadsheet evidence, whereas Agent-R$^2$-4B recovers from parsing and execution errors, computes the required values, and verifies the generated plots in the final PDF deliverable.}

    \label{fig:case_study}
\end{figure*}

\paragraph{Mitigating Shortcut Behavior Drives Improvement.}
\label{sec:shortcut_evidence}
\label{sec:shortcut_stats}
We define effective steps as tool-call steps that do not result in errors, which to some extent reflect exploration and completion depth. In Figure~\ref{fig:effective_steps_pass1}, we observe that effective steps positively correlate with GDPval Pass@1. Across model scales, agents with few effective steps tend to stop before sufficient grounding or verification, producing plausible artifacts without adequate evidence extraction or computation. We refer to this failure mode as shortcut behavior. Agent-R$^2$ mitigates this failure mode through two-stage SFT and rubric-based RL on the synthesized 4k data; in contrast, the 4B base model shows shallow interaction and low performance.

We further examine the trajectories and deliverables of the base, SFT, and RL models on all 220 GDPval tasks. Metric definitions are provided in Appendix~\ref{app:shortcut_metrics}. Table~\ref{tab:shortcut_stats} shows that both training stages reduce shortcut behavior. The base model rarely re-checks its deliverable and, in the build phase, writes the deliverable right after an unfixed error in 55\% of cases. After SFT and RL, the agent re-checks almost every deliverable, and such bypasses fall to 16\% and 11\%. Numerical errors are fixed more often at each stage, and placeholder text falls to 5.7\%, the same rate as in the expert-written GDPval deliverables. From SFT to RL, writer code that masks errors with \texttt{except}$\rightarrow$default/\texttt{pass} falls from 4.1\% to 0.9\%, and the share of traceable numbers in deliverables rises from 34.6\% to 38.8\%.
\begin{table}[t]
\centering
\small
\setlength{\tabcolsep}{3pt}
\begin{tabular*}{\linewidth}{@{\extracolsep{\fill}}p{0.55\linewidth}ccc@{}}
\toprule
\textbf{Metric (\%)} & \textbf{Base} & \textbf{SFT} & \textbf{RL} \\
\midrule
Re-checking the deliverable $\uparrow$ & 0.5 & 91.7 & \textbf{95.8} \\
Writing despite an unfixed error $\downarrow$ & 55.0 & 16.0 & \textbf{11.0} \\
Numerical-error fix rate $\uparrow$ & 60.7 & 79.6 & \textbf{85.7} \\
Deliverables with placeholder text $\downarrow$ & 12.5 & 9.6 & \textbf{5.7} \\
Traceable numbers in deliverables $\uparrow$ & 31.2 & 34.6 & \textbf{38.8} \\
\bottomrule
\end{tabular*}
\caption{Direct evidence on shortcut behavior on GDPval. All values are percentages: re-checking and placeholder rates are measured over eligible tasks, bypass and fix rates over the corresponding error episodes, and numeric traceability as the mean per-task share of traceable numbers.}
\label{tab:shortcut_stats}
\end{table}

\paragraph{Case Study on Mitigating Shortcut Behavior.}
Figure~\ref{fig:case_study} illustrates how shortcut behavior appears in a materials-test report task. The base Qwen3-4B model fails to locate the target spreadsheet values, but still writes a report with dummy evidence, producing a superficially complete deliverable. Gemini-2.5-Pro performs more inspection, yet silently converts key measurements to NaN and generates empty charts. In contrast, Agent-R$^2$-4B recovers from parsing, computation, and drawing errors, computes the required hardness values, and verifies the final plots, achieving a score far above Gemini-2.5-Pro.

\subsection{Ablation Studies}
\label{sec:ablations}

\begin{table}[t]
\centering
\small
\begin{tabular}{llr}
\toprule
\textbf{Ablation} & \textbf{Method} & \textbf{Pass@1} \\
\midrule
Data grounding & w/ synthetic files & 41.4 \\
 & w/ real files & \textbf{43.3} \\
\midrule
Reward design & w/o rubrics & 49.5 \\
 & w/ rubrics & \textbf{51.0} \\
\bottomrule
\end{tabular}
\caption{\textbf{Ablation of real files and rubric rewards on GDPval.} SFT models use matched real-file or synthetic-file tasks. RL variants use or omit rubric rewards, starting from the same SFT model.}
\label{tab:ablation}
\end{table}

To isolate the contribution of real files, we replace the attachment-searching stage with prompting the model to generate files for a 1k subset of our SFT tasks, then re-synthesize each task using the same pipeline. This yields 1k matched pairs built from the same source instances, with a similar number of attachments on both sides. Training one SFT model on each side under the same recipe yields GDPval Pass@1 of 43.3 for real files versus 41.4 for synthetic files (Table~\ref{tab:ablation}). The expert trajectories also use 2.54$\times$ more tool calls with real files (20.4 vs.\ 8.0), reflecting the layouts, structures, and cross-file dependencies that require more inspection and processing.
To isolate the contribution of fine-grained rubrics, we train an RL variant from the same SFT model without rubric rewards. Under the evaluation protocol of Section~\ref{sec:experiment_setup}, removing rubric rewards lowers GDPval Pass@1 from 51.0 to 49.5 (Table~\ref{tab:ablation}), demonstrating the benefit of rubric-based rewards.

\section{Related Work}
\label{sec:related_work}

\paragraph{Benchmarks for Real-World Productivity Tasks.}
Recent benchmarks increasingly move beyond static question answering toward realistic, long-horizon work completion. GDPval~\citep{patwardhan2025gdpvalb}, \(\$1\)M-Bench~\citep{yang2026onemillionbench}, and Agents' Last Exam~\citep{sun2026agents} evaluate economically valuable work through expert-designed occupational and high-stakes tasks. Finance Agent Benchmark~\citep{bigeard2025finance} focuses on finance research tasks grounded in SEC filings and expert rubrics. APEX--Agents~\citep{vidgen2026apexagents} and AgentSkillOS~\citep{li2026organizing} study evaluations centered on tool use and skill orchestration, where agents must select and coordinate external capabilities to complete tasks. ClawBench~\citep{zhang2026clawbench} and EdgeBench~\citep{zhu2026edgebench} study agents in real-world interactive environments, emphasizing live operations, sustained exploration, and feedback-driven progress. Workspace-Bench~\citep{tang2026workspacebench} and EnterpriseClawBench~\citep{zhong2026enterpriseclawbench} focus on workspace and enterprise settings, where agents must reason over heterogeneous files, business artifacts, and workplace-specific context.

\paragraph{Data Synthesis for Agent Post-Training.}
Synthesizing post-training data has become an important direction for improving agent capabilities. Existing work scales tool-use trajectories and executable environments from corpora, programs, or workflow templates~\citep{xu2026unlocking,song2026envscaler,xu2026envfactory,dong2026agentworld}, or replay web tutorials as verified GUI trajectories~\citep{xu2025agenttrek}. These methods are effective for scaling tool-use data, but they mainly target specific tools with predefined interfaces rather than more general tools such as terminals, which offer broader applicability for realistic productivity work. Recent data synthesis for terminal agents improves verifiability by generating command-line environments, tasks, or skills~\citep{gandhi2026endless,cheng2026terminalworld,hua2026cliuniverse}. However, these settings mainly focus on structured text files and rarely involve structurally complex office files commonly used in professional productivity work. Studies based on office files mainly focus on document-grounded question answering and analysis~\citep{zhang2026docdancera,qiao2026scaling,zhang2025deepanalyzea,chi2026spreadsheetrl} or single-type file generation~\citep{zheng2025pptagent,zheng2026deeppresenter}. In contrast, Agent-R$^2$ grounds task synthesis in real occupational workflows and authentic multi-format attachments, producing deliverable-oriented tasks together with fine-grained rubrics.

\section{Conclusion}
\label{sec:conclusion}

We introduced Agent-R$^2$, a real-source data synthesis framework that scales agents for realistic productivity work. Agent-R$^2$ constructs training instances grounded in real workflows and authentic files with fine-grained rubrics. Using 4k synthesized instances, we train Agent-R$^2$-4B with SFT and rubric-based RL. Experiments on GDPval and OfficeQA show that Agent-R$^2$-4B improves task completion and file-grounded reasoning, outperforming several larger open-source baselines. Behavioral analysis further shows that Agent-R$^2$-4B reduces shortcut behavior and improves completion reliability in complex productivity tasks.

\bibliography{aaai2027}

\clearpage
\twocolumn
\appendix
\section{Model Configuration and Training Details}
\label{app:model_training}

\subsection{Model Configuration}

Table~\ref{tab:solver_vision_models} specifies the auxiliary vision model used
by each text-only solver evaluated on GDPval. For these solvers,
\textit{view\_image} invokes the model listed in the second column and returns
its textual visual observation. To promote a fair comparison, we select, where
possible, an auxiliary vision model from the same model family and with a
parameter count comparable to that of the solver. All Agent-R$^2$ variants use
the same visual backend as their Qwen3-4B-Thinking-2507 base model. Solvers with
native visual capability are omitted from the table.

\begin{table}[h]
\centering
\scriptsize
\begin{tabular}{ll}
\toprule
\textbf{Solver model} & \textbf{Auxiliary vision model} \\
\midrule
DeepSeek-v3.1 & DeepSeek-v4-Flash \\
Qwen3-30B-A3B-Thinking-2507 & Qwen3-VL-30B-A3B-Thinking \\
Qwen3-235B-A22B-Thinking-2507 & Qwen3-VL-235B-A22B-Thinking \\
Qwen3-4B-Thinking-2507 & Qwen3-VL-4B-Thinking \\
Agent-R$^2$-4B-SFT & Qwen3-VL-4B-Thinking \\
Agent-R$^2$-4B & Qwen3-VL-4B-Thinking \\
\bottomrule
\end{tabular}
\caption{Auxiliary vision models used for text-only solvers in GDPval evaluation.}
\label{tab:solver_vision_models}
\end{table}

\subsection{Training and Evaluation Details}
\label{app:training_evaluation}

\subsubsection{Agent Runtime and Tools}
\label{app:runtime}

The agent loop is built on verl v0.6.x and runs in the open-source AIO Sandbox. Tools are exposed through the OpenAI function-calling interface. The \textit{web\_search} and \textit{web\_fetch} tools are backed by Serper; during attachment searching, queries use file-type filters such as \texttt{filetype:pptx} to target the required formats.

\subsubsection{Training Details}
\label{app:training_extra}

\begin{figure}[h]
\centering
\begin{minipage}[t]{0.48\linewidth}
\centering
\includegraphics[width=\linewidth]{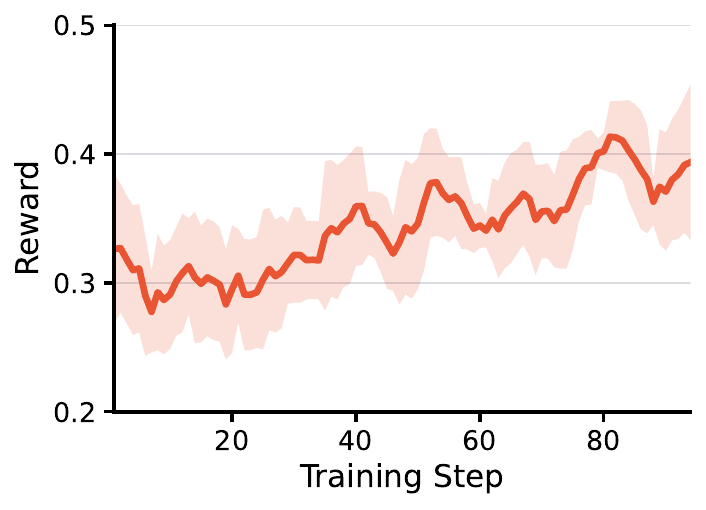}
\caption{Training reward during reinforcement learning.}
\label{fig:training_reward}
\end{minipage}
\hfill
\begin{minipage}[t]{0.48\linewidth}
\centering
\includegraphics[width=\linewidth]{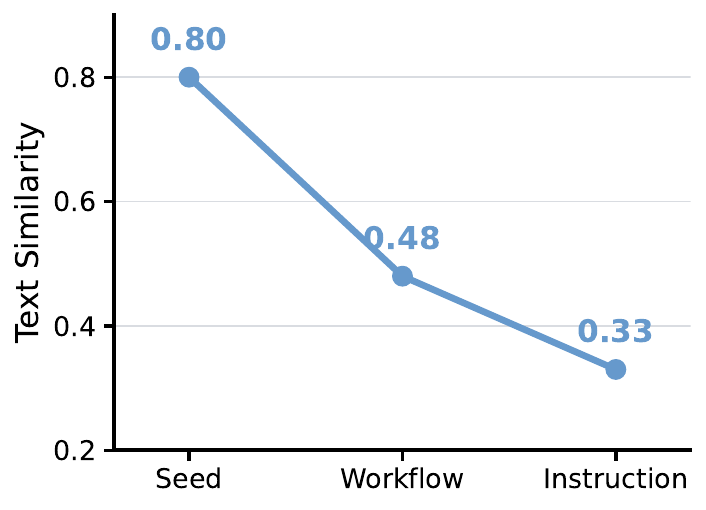}
\caption{Cosine similarity across synthesis stages.}
\label{fig:embedding_similarity}
\end{minipage}
\end{figure}

Each SFT trajectory is split into per-round samples so that every sample fits the Qwen3 chat template. SFT runs for a fixed two epochs. For both SFT and RL, we use the final checkpoint and perform no validation-based checkpoint selection; neither GDPval nor OfficeQA is used for model selection. Table~\ref{tab:rl_training_hyperparameters} summarizes the hyperparameters used for reinforcement learning. Figure~\ref{fig:training_reward} shows that the training reward follows an overall upward trend over the course of optimization.

\begin{table}[h]
\centering
\small
\begin{tabular}{cc}
\toprule
\textbf{Hyperparameter} & \textbf{Value} \\
\midrule
batch\_size & 16 \\
max\_response\_length & 65,536 \\
group\_size & 4 \\
max\_assistant\_turns & 40 \\
max\_tool\_response\_length & 4,096 \\
training\_temperature & 1.0 \\
learning\_rate & $1 \times 10^{-6}$ \\
\bottomrule
\end{tabular}
\caption{Hyperparameters used for reinforcement learning.}
\label{tab:rl_training_hyperparameters}
\end{table}

\subsubsection{Evaluation Protocol}
\label{app:eval_protocol}

All results in Tables~\ref{tab:main_gdpval_results} and~\ref{tab:officeqa} are freshly evaluated under a common runtime setup with the same sandbox and tool framework. Each run allows at most 100 steps and an unlimited number of parallel tool calls per step. API baselines use their default sampling settings, and the auxiliary vision models used by text-only solvers are listed in Table~\ref{tab:solver_vision_models}. The benchmark-specific Pass@$k$ definitions are given in the corresponding table captions. API-error retries are handled separately from the $k$ task attempts.

\begin{figure*}[!t]
    \centering
    \begin{minipage}[t]{0.31\linewidth}
        \centering
        \footnotesize
        \captionof{table}{\textbf{Attachment statistics.} Counts and complexity statistics for the four major office-oriented attachment types.}
        \label{tab:attachment_profile}
        \setlength{\tabcolsep}{3pt}
        \begin{tabular}[t]{lccc}
            \toprule
            \textbf{Type} & \textbf{Count} & \textbf{Avg Sz. (MB)} & \textbf{Mean Pg.} \\
            \midrule
            \texttt{.pdf} & 5,010 & 1.00 & 15.0 \\
            \texttt{.docx} & 3,149 & 0.25 & 13.7 \\
            \texttt{.xlsx} & 1,258 & 0.20 & -- \\
            \texttt{.pptx} & 390 & 0.78 & 23.6 \\
            \bottomrule
        \end{tabular}
    \end{minipage}
    \hfill
    \begin{minipage}[t]{0.66\linewidth}
        \centering
        \raisebox{-\height}{\includegraphics[width=\linewidth]{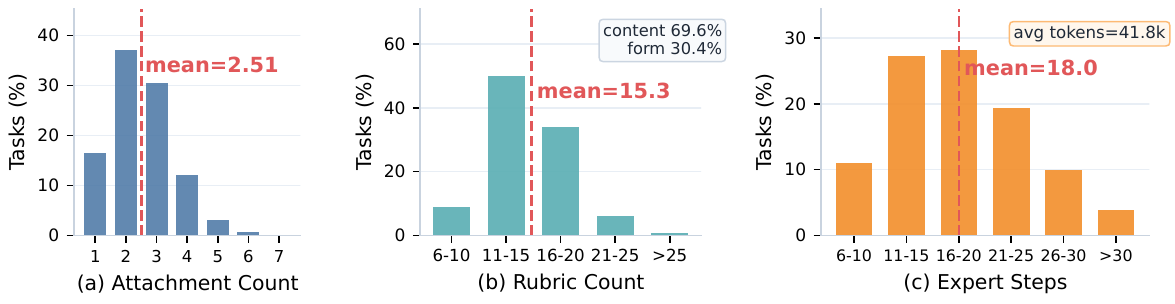}}
        \caption{\textbf{Synthesized task statistics:} attachments per task, rubric items per task, and expert-trajectory length.}
        \label{fig:data_statistics}
    \end{minipage}
\end{figure*}

\section{Additional Results}
\label{app:robustness}

\subsection{Judge Stability across Repeats and Judge Models}
\label{app:judge_stability}

GDPval scores rely on an LLM judge, so we test whether our conclusions depend on judge stochasticity or on the choice of judge. We use two judge models, Gemini-3.1-Pro and DeepSeek-v4-Flash, and each judge independently evaluates the deliverables of Agent-R$^2$-4B and Qwen3-235B-A22B-Thinking-2507 on all 220 GDPval tasks three times.

Table~\ref{tab:judge_stability} reports the range of Pass@1 across the three repeats. Per-task scores do vary between repeats, with a median absolute difference of 3.2 and 3.1 points (mean 5.7 and 5.4) under the two judges, but these differences largely cancel in the aggregate score. Agent-R$^2$-4B stays within 50.8--51.1 under Gemini-3.1-Pro, consistent with the 51.0 reported in Table~\ref{tab:main_gdpval_results}, and within 51.7--52.3 under DeepSeek-v4-Flash. The two judges differ slightly in absolute scores but agree on relative quality, with a per-task Pearson correlation of 0.93. Under both judges, the lowest score of Agent-R$^2$-4B is above the highest score of Qwen3-235B-A22B-Thinking-2507, so Agent-R$^2$-4B leads in all six evaluations.

\begin{table}[h]
\centering
\small
\setlength{\tabcolsep}{6pt}
\begin{tabular}{lcc}
\toprule
\textbf{Model} & \textbf{Gemini-3.1-Pro} & \textbf{DeepSeek-v4-Flash} \\
\midrule
Qwen3-235B-A22B & 47.7--48.2 & 49.2--49.9 \\
Agent-R$^2$-4B & \textbf{50.8--51.1} & \textbf{51.7--52.3} \\
\bottomrule
\end{tabular}
\caption{GDPval Pass@1 range over three independent evaluations by each judge model.}
\label{tab:judge_stability}
\end{table}

\subsection{Data Scaling}
\label{app:data_scaling}

To examine how performance scales with the amount of synthesized data, we train SFT models on 1k, 2k, and 4k trajectories and evaluate them on 100 held-out synthesized tasks, scored from 0 to 1 against their rubrics, as well as on GDPval. Table~\ref{tab:data_scaling} shows that scores on the held-out synthesized tasks keep rising with more data, and the win rate over the base model grows accordingly, while GDPval Pass@1 saturates. The continued gains on held-out synthesized tasks indicate that the pipeline keeps producing useful supervision as it scales, and the saturating GDPval curve shows that we do not overfit the benchmark.

\begin{table}[h]
\centering
\small
\setlength{\tabcolsep}{6pt}
\begin{tabular}{lccc}
\toprule
\textbf{SFT data} & \textbf{Held-out score} & \textbf{Win rate} & \textbf{GDPval Pass@1} \\
\midrule
0 (base) & 0.254 & -- & 32.4 \\
1k & 0.486 & 64\% & 43.3 \\
2k & 0.507 & 67\% & 44.9 \\
4k & \textbf{0.617} & \textbf{73\%} & \textbf{45.5} \\
\bottomrule
\end{tabular}
\caption{Data scaling of SFT on 100 held-out synthesized tasks and on GDPval.}
\label{tab:data_scaling}
\end{table}

\subsection{Additional Data versus the RL Stage}
\label{app:ablation_rl_vs_data}

The final model uses about 3k instances for SFT and about 1k additional tasks for RL, so its gain over the SFT model could in principle come from exposure to more data. To control for this, we train an SFT model on all 4k expert trajectories. Table~\ref{tab:ablation_rl_vs_data} shows that this 4k-SFT control reaches 45.5 Pass@1, only slightly above the 3k-SFT model (45.3) and well below SFT+RL (51.0). The gain therefore comes from RL rather than from more data, consistent with the behavioral analysis in Section~\ref{sec:analysis}.

This control also addresses whether the gains come mainly from distilling a strong teacher. Imitating all 4k teacher trajectories stops at 45.5 Pass@1, whereas RL on the same data adds 5.7 Pass@1 over the 3k-SFT model (45.3$\rightarrow$51.0).

\begin{table}[h]
\centering
\small
\setlength{\tabcolsep}{4pt}
\begin{tabular}{lcc}
\toprule
\textbf{Model} & \textbf{Training data} & \textbf{GDPval Pass@1} \\
\midrule
Qwen3-4B-Thinking-2507 & -- & 32.4 \\
Agent-R$^2$-4B-SFT & 3k SFT & 45.3 \\
4k-SFT control & 4k SFT & 45.5 \\
Agent-R$^2$-4B & 3k SFT + 1k RL & \textbf{51.0} \\
\bottomrule
\end{tabular}
\caption{Separating the effect of additional data from the RL stage.}
\label{tab:ablation_rl_vs_data}
\end{table}

\subsection{Additional Benchmark}
\label{app:generalization}

We additionally evaluate on GTA-Workflow~\citep{wang2026gta2}, which contains 132 open-ended workflows. Agent-R$^2$-4B achieves a Root Score of 4.52/10 under the same agent-as-a-judge protocol, compared with 1.53/10 for its base model and 5.15/10 for DeepSeek-v4-Flash.

\subsection{Why Effective Steps Decrease after RL}
\label{app:effective_steps}

We decompose effective steps by step type under the error definition in Appendix~\ref{app:shortcut_metrics}. Note that this definition counts only genuine error signatures, so absolute counts differ slightly from Figure~\ref{fig:effective_steps_pass1}. Table~\ref{tab:effective_step_breakdown} shows that from SFT to RL, almost the entire decrease comes from computation steps such as data processing, which drop from 11.9 to 7.3 per task. Attachment reading is unchanged, while deliverable writing and the build phase grow. As a result, RL spends a larger share of its effective steps on reading, writing, and checking the deliverable (40\%$\rightarrow$48\%).

\begin{table}[t]
\centering
\small
\setlength{\tabcolsep}{4pt}
\begin{tabular}{@{}p{0.66\linewidth}cc@{}}
\toprule
\textbf{Effective steps per task} & \textbf{SFT} & \textbf{RL} \\
\midrule
Computation & 11.9 & 7.3 \\
Attachment reading & 3.4 & 3.4 \\
Deliverable writing & 2.3 & 2.7 \\
Build phase (all step types) & 4.9 & 5.5 \\
\midrule
Share of reading, writing, and checking & 40\% & 48\% \\
\bottomrule
\end{tabular}
\caption{Decomposition of effective steps on GDPval from SFT to RL.}
\label{tab:effective_step_breakdown}
\end{table}

More steps do not mean better work. Within each model, longer trajectories score lower (per-task Spearman correlation $-0.15$ for SFT and $-0.41$ for RL), and the share of runs that hit the 100-step limit drops from 8.2\% to 3.6\%. RL thus cuts redundant computation while preserving the grounding and verification steps that matter for the deliverable. The gain comes from how the steps are used rather than how many steps are taken.

\section{Metric Definitions for Shortcut Analysis}
\label{app:shortcut_metrics}

The following definitions are used for the trajectory- and deliverable-level analysis in Section~\ref{sec:shortcut_evidence}.

All statistics are computed automatically with rule-based matching on trajectories and deliverables.
\begin{itemize}
    \item \textbf{Error.} A tool call whose output contains a genuine error signature, such as a traceback, an error message, or a missing file. Output on stderr without an error signature is not counted.
    \item \textbf{Phases.} The \emph{exploration} phase precedes the first successful deliverable write; the \emph{build} phase spans the first to the last successful write; the \emph{verification} phase follows the last successful write.
    \item \textbf{Numerical-error fix rate.} The percentage of numerical/data-logic error episodes that are subsequently fixed under the criterion above. This category includes indexing, value, type, arithmetic, and related computation exceptions.
    \item \textbf{Writing despite an unfixed error.} A deliverable is written before an error is fixed, with at most one intervening successful step. The reported rate is the percentage of build-phase error episodes that satisfy this condition.
    \item \textbf{Re-checking the deliverable.} The agent reopens or inspects the deliverable after a successful write. The rate is measured over tasks with a successful deliverable-writing operation.
    \item \textbf{Placeholder text.} The deliverable contains placeholder strings such as ``TBD'', ``[insert \ldots]'', ``lorem'', or ``placeholder''. The rate is measured over tasks with reference attachments and an output deliverable.
    \item \textbf{Error masking.} The code that writes the deliverable catches exceptions and falls back to a default value or \texttt{pass}. We report the percentage of tasks with a successful deliverable-writing operation whose writer code matches this pattern.
    \item \textbf{Traceable numbers.} The share of distinct numbers in the output deliverables that can be found in the prompt or the attachments. We compute this share for each task with reference attachments and at least five distinct output numbers, then report the mean across eligible tasks.
\end{itemize}

\section{Decontamination Audit}
\label{app:decontamination}

Our training materials are collected from web corpora and online search, so we audit the synthesized tasks and their attachments for overlap with the evaluation benchmarks at four levels.

\paragraph{File hashes and source URLs.} Neither the SHA-256 hashes of our attachment files nor the source URLs we recovered overlap with any file in GDPval or OfficeQA.

\paragraph{Source domains.} Only 0.6\% of our source domains appear in GDPval instructions, mostly broad public sources such as \texttt{irs.gov}.

\paragraph{Near-duplicate documents.} For each attachment, we compute the nearest-neighbor 5-gram MinHash Jaccard similarity to the benchmark documents. The average is 0.005 for GDPval (maximum 0.125) and 0.009 for OfficeQA (maximum 0.023). For reference, two different issues of the same U.S.\ Treasury Bulletin have a similarity of 0.36.

\paragraph{Instruction-level similarity.} For each benchmark prompt, we compute the embedding cosine similarity to its nearest training instruction. The average is 0.43 for GDPval and 0.36 for OfficeQA. For GDPval, this is below the similarity between a GDPval prompt and its nearest other GDPval prompt (0.46), and far below the similarity of a paraphrase (0.93).

\section{Data Details}
\label{app:quality_extra}
\label{app:additional_statistics_quality}

Retrieving real files does not by itself guarantee realistic tasks: topic-relevant attachments may differ in entities, dates, or assumptions, and synthesized rubrics may not be grounded in the files. This section brings together filtering decisions, synthesis statistics, manual quality audits, and data compliance and release details.

\subsection{Filtering Decisions and Data Attrition}
\label{app:filtering}

Each filtering step is implemented as a structured decision that the prompts in Appendix~\ref{app:prompts} require the model to output in JSON. For example, during workflow extraction only documents labeled \texttt{"workflow": true} are kept, and during attachment searching only files listed under \texttt{"highly\_relevant\_files"} are retained. Starting from 10k seed documents, about 2.2k are discarded at workflow extraction and about 3.8k during attachment searching and verification, leaving about 4k task instances.

\subsection{Semantic Similarity Across Synthesis Stages}

Figure~\ref{fig:embedding_similarity} reports text similarity measured by
Qwen3-Embedding-4B across synthesis stages. The final instructions have low
similarity to the seed tasks, with an average similarity of 0.33.

\subsection{Manual Quality Inspection}
\label{app:quality_audit}

We conduct a manual audit of 100 examples from the retained synthesis set across three quality dimensions: workflow--question alignment (86\%), rubric-item validity (99.8\%), and visual presentation (97\%).

Workflow and question construction are fully aligned in 86 examples, while the
remaining 14 show partial alignment. In these cases, some questions use
descriptive calculations in place of causal or simulation analysis, some omit
key cross-validation, and others omit or modify a requested artifact. Despite
these differences, all 14 examples remain topically consistent with their
workflows, although topical consistency alone does not meet the criterion for
full alignment. For rubric validity, two rubric items are inaccurate because
repeated calculations produce incorrect numerical values. For visual
presentation, the examples with room for improvement mainly contain sparse
table or continuation pages and otherwise blank trailing pages with only a
footer. These presentation issues do not affect the overall readability of the
deliverables.

\subsection{Cross-File Consistency and Task Feasibility}
\label{app:cross_file}

We further manually evaluate the same 100 tasks along two additional dimensions. \emph{Cross-file consistency} asks whether the attachments of a task describe mutually compatible entities, dates, and assumptions. \emph{Feasibility} asks whether the task can be completed from its instruction and attachments. Cross-file consistency reaches 95\%, and task feasibility reaches 85\%. Among the inconsistent tasks, one example mixes GPU specifications with data on ASIC Bitcoin miners.

\subsection{Rubric Grounding}
\label{app:rubric_grounding}

The manual audit finds that 99.8\% of rubric items are valid. To measure how strongly rubrics depend on the attachments, we trace every specific number that appears in the training rubrics. Of these numbers, 41.9\% appear in the instruction, 39.0\% appear only in the attachments, and 19.1\% appear in neither and must be computed. Most rubric numbers therefore cannot be checked from the instruction alone, so satisfying the rubric requires reading and processing the attached files.

\clearpage
\onecolumn
\section{Prompts}
\label{app:prompts}

\begin{lstlisting}[style=promptbox,caption={Workflow extraction prompt.},label={lst:scenario_prompt}]
[System]
You are a professional productivity task scenario designer. Your responsibility is to analyze web corpus content, determine whether it is suitable for designing workflow scenarios, and when suitable, mine potential workflow scenarios from it.

[Your Core Tasks]
1. Deeply analyze the input web corpus, understanding its core themes, domain background, and key information.
2. Determine whether the corpus can support one or more real, complex, file-based workflow scenarios as described in the task description.
3. If so, mine potential workflows and required files based on the corpus content and task description, conducting multi-step workflows based on one or more files.
4. You may freely combine, decompose, or derive one or more task scenarios without being limited to the preset task descriptions.

[Workflow Design Principles]
- Authenticity: Scenarios should reflect real-world work requirements that exist in practice.
- Generalizability: Workflow steps should use generic sub-goal descriptions, avoiding overly specific entity information that would be hard to find relevant files for (e.g., specific chapter numbers, codes, etc.).
- Complexity: Workflows should have sufficient complexity, requiring multi-step reasoning and operations to complete.
- Reasonableness: Required files should conform to actual work habits in the relevant domain.

[Output Format]
Please strictly follow the JSON format below.

When the corpus is **not suitable** for question design (workflow is false):
{"summary": "Brief summary of the web corpus content, describing its core theme", "workflow": false}

When the corpus is **suitable** for question design (workflow is true):
{
  "summary": "Brief summary of the web corpus content, describing its core theme",
  "workflow": true,
  "task": ["Reference task content verbatim", "Reference task content verbatim"],
  "scenario": {
    "title": "Scenario title",
    "workflow": [
      {"step": 1, "description": "Step 1 description"},
      {"step": 2, "description": "Step 2 description"}
    ]
  },
  "required_files": [
    "Description of file type and purpose",
    "Description of file type and purpose"
  ],
  "reasoning": "Rationale for designing this scenario, explaining how the web corpus supports this workflow"
}

[User]
Based on the following information, first determine whether the web corpus is suitable for question design, then decide whether to design a workflow scenario.

[Job Title]
<JOB_TITLE>

[Web Corpus Content]
<WEB_CORPUS_CONTENT>

[Reference Tasks] (Fill the task field with the verbatim content of the tasks actually used)
<CANDIDATE_OCCUPATIONAL_TASKS>

Please deeply analyze the web corpus content:
1. First determine whether the corpus is suitable for question design.
2. If suitable, mine potential workflow scenarios and fill the task field with the verbatim reference task content actually used.
3. Each item in required_files should be a single string describing the file type and purpose.

Output the result in JSON format.
\end{lstlisting}

\begin{lstlisting}[style=promptbox,caption={Attachment search and verification prompt.},label={lst:search_prompt}]
[System]
You are a professional resource search assistant. Your responsibility is to search for and retrieve the required file resources over multiple rounds based on the workflow requirements.

[Your Core Tasks]
- Analyze the types and purposes of files required by the workflow scenario, design effective search keywords, call search tools to retrieve files, evaluate whether the file content meets the scenario requirements, and rename files if their names do not conform to realistic naming conventions (e.g., Proposals%20Evaluation%20Report%20Template%20.docx).

[Search Rules]
1. Call only one search tool (web_search_*) per round. Choose the file type to search based on the workflow needs.
2. After a search tool successfully downloads a file, you must use the terminal tool or view_image tool to review the file content and evaluate whether it meets the scenario requirements before proceeding to the next round.
3. If the scenario is too specific and no results are found, try broader keywords.
4. Execute at most <MAX_STEPS> rounds; return the result once suitable files are found or the maximum number of rounds is reached.

[File Evaluation Criteria]
Evaluate file relevance across three dimensions:
1. **Topic Match** - Does the file's main subject align with the scenario's professional domain and work content?
2. **Utility Match** - Can the file's content type and data directly support the workflow tasks? (e.g., a numerical dataset supports calculations; a technical manual supports equipment assessment)

**Highly Relevant**: Satisfies all three dimensions above. The file directly provides the specific data, standards, technical specifications, or documentation that the scenario's workflow requires.

**Partially Relevant**: Satisfies 1 dimensions but with limitations, OR serves as useful background context. Common cases: the file covers a related subfield but needs significant adaptation; the file contains background knowledge but lacks task-specific details; the file was caught by a keyword hit in metadata or a cited reference rather than its main body content.

**Irrelevant**: Fails on most dimensions - the file comes from a unrelated industry, or serves a different purpose (e.g., management/process documents for a scientific research scenario, or generic intro materials for an expert-level scenario).

[Return Format]
{
  "highly_relevant_files": [
    {
      "sandbox_path": "Path of the file in the sandbox",
      "relevance": "Explanation of relevance to the scenario"
    }
  ],
  "partially_relevant_files": [
    {
      "sandbox_path": "Path of the file in the sandbox",
      "relevance": "Explanation of relevance to the scenario"
    }
  ],
  "irrelevant_files": [
    {
      "sandbox_path": "Path of the file in the sandbox",
      "reason": "Reason for not meeting requirements"
    }
  ]
}

[User]
Please search for the required files based on the following workflow scenario requirements.

[Scenario]
Scenario: <SCENARIO_TITLE>

[Workflow]
<NUMBERED_WORKFLOW_STEPS>

[Required Files Description]
<REQUIRED_FILE_TYPES_AND_PURPOSES>

[File Save Path]
Save all files to: /home/workspace/search_files

[Finalization Message]
Based on the files already searched and inspected, return the result in JSON format.
\end{lstlisting}

\begin{lstlisting}[style=promptbox,caption={Task question and rubric generation prompt.},label={lst:question_prompt}]
[System]
You are a professional productivity task question designer. Based on workflow scenarios and retrieved files, design high-quality productivity questions and generate rubric scoring criteria.

**File System**
- `/mnt/user-data/uploads` -- Relevant files have been uploaded here. Read directly from this path.
  - Each file may have a companion **preprocessed folder** (same name as the file, without extension) containing `content.md` and extracted images. These are provided **only to help you understand the file content** -- do NOT treat them as attachments for the question.
- `/home/workspace` -- Scratch space for temp files and intermediate work. Not visible externally.
- `/mnt/user-data/outputs` -- Copy selected **original files** (not preprocessed folders) from `/mnt/user-data/uploads` here as question attachments. Don't include other files.

[Your Core Tasks]
1. Thoroughly understand the content of the provided files.
2. Based on the workflow and file content, design a question that requires producing a real deliverable.
4. Generate verifiable rubric scoring criteria.

[Question Design Principles]
- You must analyze the files in detail before designing questions. Do not skip file analysis and jump directly to question design.
- Questions should reflect real-world work requirements, require multi-step reasoning and operations, and be closely related to the attachment content.
- Deliverable types: pdf, pptx, xlsx. Choose the deliverable type that best fits the task nature.
- Not all provided files need to be included, you can choose the most suitable files for the question.
- You can adjust and enrich the workflow appropriately based on the file content.
- **Keep the question description concise**: do NOT embed all details from the attachments into the question text. The question should set the task goal and context; solvers are expected to open the attachments and infer the specifics themselves.
- Questions should ideally test the model's ability to extract information from complex file structures (e.g., nested tables, multi-sheet workbooks, structured spreadsheets, or hierarchically organized documents).

[Rubric Rules]
1. **Source-backed and verifiable**: Each criterion must clearly determine "satisfied/not satisfied"; the basis must come from the question description or attachment content. Do not fabricate requirements or values without a source.
2. **Content category** (category=content): Assesses information retrieval, multi-step reasoning, and specific operations required to complete the task, e.g., extracting key data from attachments, calculating or summarizing according to specified logic, executing specific analysis steps.
3. **Form category** (category=form): Assesses the structural and formatting requirements of the deliverable, e.g., file format, section structure, number of charts, naming conventions.

[Return Format]
After completing file analysis, first copy the selected attachments from /mnt/user-data/uploads to /mnt/user-data/outputs, then return the following JSON:
{
  "question": {
    "title": "Question title",
    "description": "Detailed question description (background context, task requirements, delivery requirements)",
    "deliverable": {
      "type": "pdf|pptx|xlsx|docx",
      "filename": "Suggested filename"
    },
    "attachments": [
      {
        "filename": "Attachment filename (already copied to outputs directory)",
        "reason": "Reason for selecting this attachment"
      }
    ]
  },
  "rubric": [
    {
      "criterion": "Specific verifiable scoring description",
      "category": "content|form",
      "points": 2,
      "source": "Source basis for the rubric and the approach needed to fulfill it"
    }
  ]
}

You have at most <MAX_STEPS> steps. Return the result when your task is complete.

[User]
Please design a productivity task question and generate rubric scoring criteria based on the following workflow scenario and retrieved files.

[Job Title] <JOB_TITLE>

[Scenario Description]
Scenario title: <SCENARIO_TITLE>

[Workflow Description]
<NUMBERED_WORKFLOW_STEPS>

[Relevant Files] (Uploaded to /mnt/user-data/uploads, annotated with relevance type)
Note: Preprocessed folders are for reference only to help you understand file content. Only original files may be used as question attachments.
<FILE_PATHS_RELEVANCE_AND_PREPROCESSED_FOLDERS>

Please start analyzing the files and designing the question, then return the result in JSON format.
\end{lstlisting}

\begin{lstlisting}[style=promptbox,caption={Expert solving prompt.},label={lst:solver_prompt}]
[System]
You are a capable, resourceful AI Agent. You cut through ambiguity, plan before acting, and deliver high-quality results.

**File System**
- `/mnt/user-data/uploads` - User-uploaded files. Read directly from this path.
- `/home/workspace` - Scratch space for temp files and intermediate work. Not visible to the user.
- `/mnt/user-data/outputs` - Save final deliverables here before replying. Files not copied here are invisible to the user.

You have at most <MAX_STEPS> steps to complete the task.

[User]
Please complete the following work task.

[Task Title]
<TASK_TITLE>

[Task Description]
<TASK_INSTRUCTION>

[Deliverable]
- File type: <DELIVERABLE_TYPE>
- Filename: <DELIVERABLE_FILENAME>

[Input Attachments]
<ATTACHMENT_PATHS_AND_DESCRIPTIONS>
Input attachments have been uploaded to the /mnt/user-data/uploads/ directory.

Please first review the content of the input attachments, then complete the task as required. Save the output file to the /mnt/user-data/outputs/ directory.
\end{lstlisting}

\section{Seed Task Example: Lawyers}
\label{app:lawyer_seed_tasks}

As an example of the seed task source, we list the task statements for the
Lawyers occupation from the O*NET seed task file. The occupation code is
23-1011.00.

\small
\begin{enumerate}
\item 3774, Core. Interpret laws, rulings and regulations for
individuals and businesses.
\item 3775, Core. Analyze the probable outcomes of cases, using
knowledge of legal precedents.
\item 3778, Core. Gather evidence to formulate defense or to initiate
legal actions by such means as interviewing clients and witnesses to ascertain
the facts of a case.
\item 3779, Core. Represent clients in court or before government
agencies.
\item 3777, Core. Evaluate findings and develop strategies and
arguments in preparation for presentation of cases.
\item 3773, Core. Advise clients concerning business transactions,
claim liability, advisability of prosecuting or defending lawsuits, or legal
rights and obligations.
\item 3780, Core. Examine legal data to determine advisability of
defending or prosecuting lawsuit.
\item 20877, Core. Prepare, draft, and review legal documents, such as
wills, deeds, patent applications, mortgages, leases, and contracts.
\item 3783, Core. Study Constitution, statutes, decisions,
regulations, and ordinances of quasi-judicial bodies to determine ramifications
for cases.
\item 3786, Core. Negotiate settlements of civil disputes.
\item 3789, Core. Supervise legal assistants.
\item 23992, Core. Negotiate contractual agreements.
\item 3787, Core. Confer with colleagues with specialties in
appropriate areas of legal issue to establish and verify bases for legal
proceedings.
\item 3788, Core. Search for and examine public and other legal
records to write opinions or establish ownership.
\item 3790, Core. Perform administrative and management functions
related to the practice of law.
\item 3776, Supplemental. Present and summarize cases to judges and
juries.
\item 3781, Supplemental. Select jurors, argue motions, meet with
judges, and question witnesses during the course of a trial.
\item 3782, Supplemental. Present evidence to defend clients or
prosecute defendants in criminal or civil litigation.
\item 3792, Supplemental. Probate wills and represent and advise
executors and administrators of estates.
\item 3785, Supplemental. Prepare legal briefs and opinions, and file
appeals in state and federal courts of appeal.
\item 3791, Supplemental. Act as agent, trustee, guardian, or executor
for businesses or individuals.
\item 3793, Supplemental. Help develop federal and state programs,
draft and interpret laws and legislation, and establish enforcement procedures.
\end{enumerate}
\normalsize

\section{Example}
\label{app:generated_example_demo}

We include one generated example from the synthesis pipeline. The generated
task is titled ``Election Contest Legal Analysis and Litigation
Strategy Memorandum.'' The original corpus for this example comes from the
first sample in the DESIGNER sample set~\citep{liu2025designer}, following the
corpus source described in Section~\ref{sec:data_synthesis}.

\begin{lstlisting}[style=promptbox,caption={Example workflow.},label={lst:generated_demo_workflow}]
[Workflow]
1. Research and interpret relevant election contest statutes and regulations.
2. Examine election records and certification documents to identify potential grounds for contesting the results.
3. Analyze legal precedents from similar contested elections.
4. Evaluate legal arguments, including equal protection claims and Voting Rights Act challenges.
5. Develop a litigation strategy and prepare legal briefs for potential state and federal court proceedings.
\end{lstlisting}

\begin{lstlisting}[style=promptbox,caption={Example question description.},label={lst:generated_demo_question}]
Background Context:
You are a senior attorney at a law firm specializing in election law. Your client, a losing candidate in a recent Virginia General Assembly election, is considering contesting the election results. The certified results show your client lost by 0.8% of the total votes cast. Additionally, there are concerns about mail-in ballots that were received after election day but postmarked before election day being counted, similar to the issue in Bost v. Illinois State Bd. of Elections.

Task Requirements:
Prepare a comprehensive legal memorandum analyzing the viability of an election contest and developing a litigation strategy. Your memorandum must:

1. Analyze the Virginia recount and contest statutes to determine:
   - Whether your client is eligible to request a recount based on the margin of victory
   - The specific filing deadlines and procedural requirements for initiating a recount petition
   - The appropriate court venue for filing the petition
   - The bond requirements for contesting a General Assembly election

2. Apply the precedent from Bost v. Illinois State Bd. of Elections (2026) to assess:
   - Whether your client has standing to challenge the counting of mail-in ballots received after election day
   - The legal arguments supporting standing based on the Supreme Court's majority opinion
   - Potential counterarguments based on the dissenting opinion
   - How the concurrence's pocketbook injury theory might strengthen or weaken your position

3. Develop a litigation strategy that addresses:
   - The sequence of legal actions (recount petition followed by contest, or direct contest)
   - Specific grounds for contest under Virginia law that would have probable impact on the election outcome
   - Timeline for all required filings and proceedings
   - Evidence preservation and discovery considerations

Delivery Requirements:
Prepare a formal legal memorandum in PDF or DOCX format with the following sections:
- Executive Summary (maximum 1 page)
- Legal Framework Analysis (Virginia recount and contest statutes)
- Standing Analysis (application of Bost precedent)
- Litigation Strategy and Recommendations
- Timeline and Procedural Checklist
- Conclusion

The memorandum should be professionally formatted with proper legal citations, clear headings, and actionable recommendations.
\end{lstlisting}

\begin{lstlisting}[style=promptbox,caption={Example rubric.},label={lst:generated_demo_rubric}]
[Rubric]
1. criterion: Correctly identifies that a recount petition is permissible because the 0.8% margin is within the 1% threshold specified in Virginia law
   category: content
   source: Virginia Recounts and Contests document states: "A petition for a recount may be filed only if there is a difference of not more than one percent (1%) of the total vote cast"

2. criterion: Accurately states the 10-day filing deadline for recount petitions from the date of certification
   category: content
   source: Virginia document states: "a petition for a recount of an election, other than an election for presidential electors, shall be filed within 10 days from the day the State Board or the electoral board certifies the result"

3. criterion: Identifies the correct venue for filing a General Assembly election contest petition (circuit court where the challenged candidate resides)
   category: content
   source: Virginia document states: "For non-statewide offices, the petition must be filed in the circuit court of the county or city in which the candidate being challenged resides"

4. criterion: Correctly identifies the bond requirement for contesting a General Assembly election as $100 per precinct
   category: content
   source: Virginia document states: "the contestant must post a surety bond in the amount of $100 for each precinct contained in whole or in part of the district contested"

5. criterion: Accurately explains the standing holding from Bost v. Illinois State Bd. of Elections that candidates have standing to challenge vote-counting rules based on their interest in a fair electoral process
   category: content
   source: Bost case syllabus states: "As a candidate for office, Congressman Bost has standing to challenge the rules that govern the counting of votes in his election"

6. criterion: Correctly identifies that under Bost, candidates do not need to show a substantial risk of election loss to establish standing
   category: content
   source: Bost majority opinion states: "Candidates do not need to show a substantial risk that a rule will cause them to lose the election or prevent them from achieving a legally significant vote threshold in order to have standing"

7. criterion: Identifies and explains the reputational harm theory from Bost as a basis for standing (erosion of public confidence undermines political legitimacy)
   category: content
   source: Bost opinion states: "Rules that undermine the integrity of the electoral process also undermine the winner's political legitimacy" and "reputational harms are classic Article III injuries"

8. criterion: Addresses the dissenting opinion's counterargument that the interest in fair elections is a generalized grievance common to all voters
   category: content
   source: Bost dissent states: "the interest in a fair electoral process is common to all members of the voting public" and candidates cannot sue based only on "generalized grievances about the conduct of government"

9. criterion: Explains Justice Barrett's concurrence regarding pocketbook injury and mitigation costs as an alternative standing theory
   category: content
   source: Bost concurrence states: "Congressman Bost has standing because he has suffered a traditional pocketbook injury" from costs incurred "to mitigate or avoid the substantial risk of harm"

10. criterion: Correctly identifies the notice deadline for contesting a General Assembly election (30 days after election or 3 days after recount conclusion, whichever is later)
    category: content
    source: Virginia document states: "Notice must be submitted within thirty days following the date of the election or three days after the conclusion of a recount, whichever is later"

11. criterion: Identifies the required grounds for contest (eligibility objections or conduct/results objections with probable impact on outcome)
    category: content
    source: Virginia document states grounds must include "Objections to the contestee's eligibility based on specific allegations" or "Objections to the conduct or results of the election, accompanied by specific allegations that, if proven true, would have a probable impact on the outcome"

12. criterion: Provides a logically sequenced litigation strategy addressing whether to pursue recount first or proceed directly to contest
    category: content
    source: Requires multi-step reasoning based on Virginia document provisions about recount procedures (Section 15.2) and contest procedures (Section 15.3), including the provision that a candidate who loses after recount may contest

13. criterion: Includes a timeline or procedural checklist covering all key deadlines (recount petition, preliminary hearing, contest notice, contestee answer, depositions)
    category: content
    source: Requires synthesis of multiple deadlines from Virginia document: 10 days for recount petition, 5-7 days for preliminary hearing, 30 days/3 days for contest notice, 10 days for contestee answer, 20-30 days for depositions

14. criterion: Memorandum includes all six required sections (Executive Summary, Legal Framework Analysis, Standing Analysis, Litigation Strategy, Timeline/Checklist, Conclusion)
    category: form
    source: Question delivery requirements specify six required sections

15. criterion: Document is properly formatted as a legal memorandum with clear headings, professional formatting, and proper legal citations
    category: form
    source: Question delivery requirements specify "formally formatted with proper legal citations, clear headings, and actionable recommendations"

16. criterion: Executive Summary is present and does not exceed 1 page
    category: form
    source: Question delivery requirements specify "Executive Summary (maximum 1 page)"

17. criterion: Legal citations reference both the Virginia Code sections and the Bost case with appropriate pinpoint citations
    category: form
    source: Question requires "proper legal citations" and attachments contain specific Code sections (Sections 24.2-800 through 24.2-814) and Bost case citations

18. criterion: Provides actionable recommendations rather than merely summarizing legal principles
    category: content
    source: Question delivery requirements specify "actionable recommendations"
\end{lstlisting}

The source workflow starts from a legal scenario involving a contested Virginia
Attorney General election and asks the model to move from statutory
interpretation and record review to precedent analysis, argument evaluation,
and litigation planning. The generated question instantiates this workflow as a
solver-facing task: it asks for a comprehensive legal memorandum with six
required sections, including an executive summary, legal framework analysis,
standing analysis, litigation strategy and recommendations, a timeline and
procedural checklist, and a conclusion. The substantive task requires the
solver to analyze Virginia recount and contest statutes, apply \textit{Bost v.
Illinois State Board of Elections} to candidate standing for challenging
late-counted mail-in ballots, and recommend a litigation sequence covering
recount, contest, evidence preservation, and discovery considerations.

The rubric evaluates whether the response identifies the 1\% recount
threshold, the 10-day recount petition deadline, the correct venue for a
General Assembly contest, the bond requirement, and the contest notice
deadline. It also checks whether the response explains the majority,
dissenting, and concurring standing theories in \textit{Bost}, provides a
coherent litigation strategy, includes the requested memorandum sections, uses
proper legal citations, and gives actionable recommendations rather than only
summarizing legal principles.


\end{document}